\documentclass[review]{elsarticle}

\usepackage{url}
\usepackage{float}
\usepackage{xcolor}
\usepackage[numbers]{natbib}
\usepackage{commath}
\usepackage{placeins}
\usepackage{booktabs}
\usepackage{multirow}
\usepackage{multicol}
\usepackage{graphicx}
\usepackage{amssymb}
\newcommand{\iu}{{i\mkern1mu}}

\usepackage{changes}

\usepackage{lineno,hyperref}
\modulolinenumbers[5]

\journal{Pattern Recognition}

\begin{document}

\begin{frontmatter}

\title{A Deeper Look into Convolutions via Eigenvalue-based Pruning}


\author[mymainaddress]{Ilke Cugu}

\author[mymainaddress]{Emre Akbas\corref{mycorrespondingauthor}}
\cortext[mycorrespondingauthor]{Corresponding author}
\ead{emre@ceng.metu.edu.tr}

\address[mymainaddress]{Department of Computer Engineering, Middle East Technical University, 06800 Ankara, Turkey}

\begin{abstract}
Convolutional neural networks (CNNs) are able to attain better visual recognition performance than fully connected neural networks despite having much fewer parameters due to their parameter sharing principle. Modern architectures usually contain a  small number of fully-connected layers, often at the end, after multiple layers of convolutions. In some cases, most of the convolutions can be eliminated without suffering any loss in recognition performance. However, there is no solid recipe to detect the hidden subset of convolutional neurons that is responsible for the majority of the recognition work. In this work, we formulate this as a pruning problem where the aim is to prune as many kernels as possible while preserving the vanilla generalization performance. To this end, we use the matrix characteristics based on eigenvalues for pruning, in comparison to the average absolute weight of a kernel which is the de facto standard in the literature to assess the importance of an individual convolutional kernel, to shed light on the internal mechanisms of a widely used family of CNNs, namely residual neural networks (ResNets), for the image classification problem using CIFAR-10, CIFAR-100 and Tiny ImageNet datasets.
\end{abstract}

\begin{keyword}
pruning \sep model compression \sep convolutional neural networks \sep deep learning \sep expert kernels \sep eigenvalues
%\texttt{elsarticle.cls}\sep \LaTeX\sep Elsevier \sep template
%\MSC[2010] 00-01\sep  99-00
\end{keyword}

\end{frontmatter}

\section{Introduction}
\label{sec:intro}

Modern deep neural networks are able to attain impressive performance on computer vision problems by using millions of parameters scattered across multiple layers in the form of convolutions. However, as we show in this study, it is possible to eliminate most of these parameters without any negative effect on the generalization error. A classical method of doing so is to prune the parameters that are effectively zero. However, in the case of pruning \emph{groups of parameters} such as the channels in a convolutional neuron, it is not clear how to assess their importance.

In the literature, the de facto standard of assessing neuron importance is to check the average absolute value of its weights, which is an effective criteria for fully connected neural nets \cite{han2015learning,han2015deep,wen2016learning,guo2016dynamic,li2016pruning,szegedy2016rethinking,yang2017designing,son2018clustering,ye2019adversarial,blakeney2020pruning}. However, for visual recognition, convolutional neural nets (CNNs) \cite{fukushima1982neocognitron,lecun1989backpropagation} usually outperform fully connected nets in terms of both accuracy and efficiency. CNNs reduce the number of parameters dramatically due to their parameter sharing principle and local connectivity. Although the general tendency is to keep several fully connected (FC) layers right before the output layers \cite{he2016deep,szegedy2016rethinking,howard2017mobilenets}, it has been shown that one can get rid of the FC layers entirely while preserving competitive performance \cite{long2015fully,iandola2016squeezenet,sandler2018mobilenetv2}. Today, the modern visual recognition architectures mostly consist of convolutions. Hence, we can think of convolutions as a very effective model compression method in and of itself. However, the story does not end there, since not all kernels are equally important for recognition. In the following, we refer to a channel in a convolutional neuron as \emph{kernel}. Suppose the input tensor is $m \times m \times c$ dimensional. A convolutional neuron defined on this input would have $k \times k \times c$ parameters. We call each of these $k \times k$ matrices as a kernel.

As we also show here, depending on the \emph{difficulty} of a problem, it is possible that only a small portion of the kernels are actually vital for visual recognition \cite{li2018measuring}. When the absence of a kernel results in a significant rise in the generalization error, we call that kernel an \textbf{expert kernel}, and emergence of expert kernels are often enforced via \emph{regularization} which is also used to limit the capacity for memorization \cite{zhang2016understanding}. However, several studies on the generalization properties of deep learning state that the convolutional case is more difficult to analyze \cite{li2018measuring,arora2018stronger}, so the theoretical advances mostly stay in the fully connected realm.

In this work, we study convolutions and ask the question: ``How can we characterize or identify expert kernels?". We formulate this question as a pruning problem, and focus on the fact that kernels are basically tiny matrices applied on a given channel of an input. Thus, we employ \textbf{matrix characteristics} as heuristics to assess the significance of a given kernel. In the model compression literature (Section \ref{sec:related_works}), this motivation puts our work under the category of \textbf{unstructured heuristic-based kernel-level pruning}. We aim to improve our understanding about CNN pruning heuristics, including eigenvalue-based ones which were not explored before. We are interested in analyzing the pruning heuristics which maintain the recognition performance (of the unpruned model) and do not require additional training after the pruning.

Discovering the correct recipe to recognize the expert kernels may influence both the theory and practice. Regarding the theory side, there are exciting empirical observations on \textbf{overparameterization,} such as the lottery ticket hypothesis \cite{frankle2018lottery} and the double-descent in generalization error \cite{belkin2019reconciling,nakkiran2019deep}. These studies try to explain deep neural nets' ability to avoid overfitting despite having much more parameters than the number of available samples, and we argue that the definition of expert kernels can shed new light on the discussion on overparameterization in deep learning. In practice, elimination of non-expert kernels may produce very sparse models. As we show in this study, the sparsity structure of the pruned model heavily depends on the problem definition (i.e., the dataset), hence, in some cases, it is possible to remove large groups of neurons resulting in a direct drop in the computational cost. Nevertheless, such models when combined with the advances in high performance sparse matrix computations can yield families of energy efficient neural nets \cite{schwartz2020green}. 

In this paper, we present, to the best of our knowledge, \textbf{the first extensive analysis on eigenvalue based heuristics to mark the expert kernels, the distribution of expert kernels through layers, and the change in their pruning behavior during different phases of training} for image classification problems using CIFAR-10/100 \cite{krizhevsky2009learning}, and Tiny ImageNet \cite{tinyimagenet} datasets. Specifically, we found that (1) for identifying expert kernels, spectral norm is a more robust heuristic -- in the sense that it better preserves generalization accuracy -- than the widely-used average absolute weight of a kernel. (2) We found that, in some cases, pruning by looking only at the real part of the largest absolute eigenvalue can be used to push the pruning ratio further without a significant loss in recognition performance by eliminating the low-performing expert kernels. (3) The dataset used in training has a distinct footprint (i.e. pattern) on the distribution of active parameters across layers after pruning. (4) The gap between the largest and the smallest absolute eigenvalues increases after the first drop in learning rate, and this causes some of the heuristics to fail in identifying expert kernels.

This paper is organized as follows: we start by reviewing the relevant work in the literature in Section \ref{sec:related_works}. Next, we introduce our candidate heuristics to assess kernel quality in Section \ref{sec:pruning_heuristics}, and the relevant regularization in Section \ref{sec:regularization}. Then, our analyses follow. Specifically, (1) we compare recognition performances of pruning heuristics in Section \ref{sec:performance}; (2) since most of our heuristics are based on eigenvalues, we discuss real vs. complex eigenvalues in Section \ref{sec:eig_analysis}; (3) we present an analysis on which heuristic prunes which kernels in Section \ref{sec:set_analysis}; (4) we present an analysis on layer-wise pruning  in Section \ref{sec:pruning_per_layer}, and (5) we present an analysis on epoch-wise pruning in Section \ref{sec:pruning_through_epochs}.

\section{Related Work}
\label{sec:related_works}

There are several ways of model compression in deep learning such as quantization, binarization, low-rank approximation, knowledge distillation, architectural design and pruning. In this work, we are only interested in pruning since it is the only category that directly targets the \textbf{inessential} parts of deep neural nets whereas the closest family of compression techniques in this regard focuses on approximating the \textbf{overall computation} (low-rank approximation).

In the literature, some of the early examples of pruning research are done on saliency based methods \cite{mozer1989skeletonization,lecun1990optimal,mpitsos1992convergence,hassibi1993second}, and the interest continues as new deeper architectures are being introduced \cite{Laine2017,dong2017learning,yu2018nisp}. In this line of work, the importance of a unit (weight/neuron/group) is determined by its direct effect on the training error. Although the required computation is immense, and hence the emphasis is on the development of approximators, the reasoning behind this approach is solid and straightforward.

Another line of work does pruning based on unit outputs. To our knowledge, the first study \cite{sietsma1988neural} of this kind looks for invariants. These invariants are defined as units whose outputs do not change across different input patterns. In other words, the units that act like bias terms are removed to save space. This is not a widely used technique anymore, perhaps it is because the large datasets of today make it hard for units to be invariant. Today, (1) the focus is on minimizing the reconstruction error of output feature maps per layer \cite{he2017channel,luo2017thinet}, (2) clustering units w.r.t to their activations \cite{dubey2018coreset,baykal2018data,mussay2019data}. We can also count the latter studies under another category, namely clustering-based pruning \cite{nowlan1992simplifying,ullrich2017soft,son2018clustering,li2019learning}. A relatively similar study \cite{he2019filter} uses the geometric median to set representative/centroid filters and prune the ones that are close to the geometric median. Moreover, recently, model explanation methods such as layer-wise relevance propagation \cite{bach2015pixel} and activation maximization \cite{zeiler2014visualizing} are also employed for neural-activation based pruning \cite{yeom2021pruning,yao2021deep}.

Next set of studies \cite{liu2017learning,ye2018rethinking,gordon2018morphnet,zhao2019variational} leverage scaling factors such as the $\gamma$ of batch norm \cite{ioffe2015batch} or introduce their own for different levels of analysis (neuron/group/layer) \cite{huang2018data} to cut off the outputs of some units. The advantage of this method is that it releases the input dependency of activation based importance assignment. In other words, if the scaling factor is $0$, it is guaranteed that the corresponding unit has no contribution to recognition. We can include mask-based pruning methods \cite{junjie2020dynamic,ramakrishnan2020differentiable} under this category as well since the basic principle is the same. Although quite useful in practice, this approach does not tell us anything about the nature of a ``good unit" for recognition. The decision is left entirely to the optimization by simply adding a regularization term for the scaling factors. Similarly, a recent work \cite{lin2019towards} also leaves the importance assignment to the learning itself by employing a minimax game as in generative adversarial networks \cite{goodfellow2014generative}. One way to study this black box is to introduce heuristics and test their validity. A well-known heuristic is the average of the absolute value of the weights of a unit \cite{han2015learning,han2015deep,wen2016learning,guo2016dynamic}. Hence, we take it as a baseline for our matrix characteristics based heuristics (see Section \ref{sec:pruning_heuristics}).

There are also analysis papers that can be found related to ours since the motivation is the same: to understand deep convolutional neural nets via pruning. Liu et al.'s work \cite{liu2018rethinking} presents an extensive empirical study on pruning algorithms to see if there is indeed an advantage of pruning over training the target architecture from scratch. Their methodology involves a fine-tuning phase after pruning to recover the loss in performance. In that setting, it is shown that \textbf{structured pruning} can be omitted, and, instead, one can directly train the target architecture from scratch and achieve competitive performance. However, it seems this is not the case for \textbf{unstructured pruning}. In our study, we also employ an unstructured pruning scheme, but we do not have a fine-tuning phase since our aim is to prune as much as possible while preserving the vanilla performance. On another note, pruning is argued to be important since one cannot achieve the same adversarial robustness or performance by training from scratch \cite{ye2019adversarial}. In addition, a recent study \cite{blakeney2020pruning} confirms the observations of Liu et al. \cite{liu2018rethinking} and studies the differences between the representations of the pruned nets and vanilla nets.

\section{Method}
In this section, we first introduce the pruning method and the heuristics that we use to identify expert kernels. Then, we describe the regularization method we used in training the neural networks.

% \subsection{Pruning Method and Heuristics}
% \label{sec:compression_modes}

Convolutional layers are quite interesting since they can express more representational power than the fully-connected counterparts despite having much fewer parameters. However, in the pruning literature, convolutions are often treated as if they are fully connected. A common approach is to compute the average absolute weight of a given kernel. However, it is not clear if it is a good indicator of an expert kernel. 

Here let us remember the definition of \emph{kernel} for convenience. Suppose the input tensor to a convolutional neuron is $m \times m \times c$ dimensional. The convolutional neuron defined on this input would have $k \times k \times c$ parameters. We call each of these $k \times k$ matrices as a kernel. When the absence of a kernel results in a significant rise in the generalization error, we call that kernel an \emph{expert kernel}.

\subsection{Pruning Method}
To decide whether to prune (i.e. remove) a  kernel, we compute the value of a \emph{pruning heuristic} on that kernel. The pruning heuristic is a measure that is computed only on the parameters on of the kernel; it does not depend on the input or output of the kernel/neuron.  The list of pruning heuristics are given below in the following section. If the value of the pruning heuristic is lower than a significance threshold, then we prune (i.e. remove) that kernel. Significance thresholds for different pruning heuristics are different. We mention them when we describe each heuristic below. Detailed information can be found in Section 1.4 of the Supplementary Material.

\subsection{Pruning Heuristics}
\label{sec:pruning_heuristics}
We propose several measures to be computed on a given kernel. The value of this measure is used as the criterion for pruning (see the Pruning Method above). We call these mesaures as \emph{pruning heuristics}. Below we describe each pruning heuristic defined for a $k \times k$ real kernel $K$.

Note that since the exact calculation of the eigenvalues of a matrix larger than $4 \times 4$ is, in general, not possible, according to the insolvability theory of Abel \cite{abel1826beweis} and Galois \cite{galois2011mathematical}, some of the pruning heuristics are only applicable to a restricted set of convolutional architectures. On the other hand, most well-known models in the literature excessively use $3 \times 3$ kernels. Moreover, at this size, the difference between the time it takes to compute different heuristics is negligible.

\subsubsection{\textbf{det}: Absolute value of the determinant}
This pruning heuristic is formally defined as 

\begin{equation}
 \abs{\det(K)}.
\end{equation}

If we assume values smaller than $10^{-12}$ as effectively 0, then this heuristic prunes kernels that are not full-rank. Therefore, its corresponding hypothesis is that an expert kernel is a full-rank matrix.

\subsubsection{\textbf{det\_gram}: Absolute value of the determinant of the Gramian matrix}
This pruning heuristic is formally defined as 

\begin{equation}
     \abs{\det(G)}
\end{equation}
where $G = K^T K$.
    
Since we employ weight regularization (see Section \ref{sec:regularization}), kernels tend to have very small weights even if they are not pruned. Combined with a threshold value of $10^{-24}$, this heuristic checks the possible numerical instability caused by weight regularization. Formally,
    
\begin{equation}
    \det(G) = \det(K^T K) = \det(K^T)\det(K) = \det(K)^2
\end{equation}
which is basically the same criteria as det. However, in Section \ref{sec:set_analysis}, we will see that the floating point constraints prevent the two pruning heuristics from being identical.

\subsubsection{\textbf{min\_eig}: The smallest absolute eigenvalue}
This pruning heuristic is formally defined as 

\begin{equation}
    \min{ \{ \abs{\lambda_i (K)} | i \in \{ 1,...,k \} \} }
\end{equation}

This heuristic basically tests the hypothesis of det heuristic since the smallest absolute eigenvalue is enough to check whether a matrix is full-rank. However, det is not enough for that due to the fact that it is the product of all eigenvalues, in which case the largest absolute eigenvalue may prevent an overkill.

\subsubsection{\textbf{min\_eig\_real}: The real part of the smallest absolute eigenvalue}
This pruning heuristic is formally defined as

\begin{equation}
    \min{ \{ \abs{a_i (K)} | i \in \{ 1,...,k \} \}}
\end{equation}
where $\lambda = a + b \iu$. In this work, we are dealing with $3 \times 3$ kernels, and these matrices can have complex conjugate eigenvalues. Hence, this heuristic enables us to measure the relevance of the imaginary part of such eigenvalues.

\subsubsection{\textbf{spectral\_radius}: The largest absolute eigenvalue}
This pruning heuristic is formally defined as

\begin{equation}
     \max{ \{ \abs{\lambda_i  (K)} | i \in \{ 1,...,k \} \} }
\end{equation}
    
This heuristic measures the largest scaling of the eigenvectors of $K$ while being a safe comparison factor for a possible overkill due to min\_eig.

\subsubsection{\textbf{spectral\_radius\_real}: The real part of the largest absolute eigenvalue}
This pruning heuristic is formally defined as

\begin{equation}
    \max{ \{ \abs{a_i (K)} | i \in \{ 1,...,k \} \} }
\end{equation}
where $\lambda = a + b \iu$. It has the same reasoning with min\_eig\_real.
    
\subsubsection{\textbf{spectral\_norm}: The largest singular value}
This pruning heuristic is formally defined as    
    
\begin{equation}
    \sqrt{\lambda_{max} (G)}
\end{equation}
where $G = K^H K = K^T K$ since the kernel $K$ is a real matrix. This heuristic is quite interesting since it is the key component of residual network invertibility \cite{behrmann2019invertible} and stable discriminator training in GANs \cite{miyato2018spectral}. Hence, naturally, we also want to test the applicability of spectral norm as an importance indicator for pruning convolutional kernels. Notice that $G$ is a $k \times k$ real symmetric matrix, so $\lambda_i (G) \in \mathbb{R}$ and $\lambda_i (G) \geq 0$  $\forall i \in \{1,..,k\}$. Thus, there is no spectral\_norm\_real heuristic.
    
\subsubsection{\textbf{weight}: The average of absolute weights}
This pruning heuristic is formally defined as 

\begin{equation}
    \frac{1}{k^2} \sum_i^k \sum_j^k \abs{w_{ij}}
\end{equation}
where $w_{ij} \in K$. This is the control group since it is the de-facto standard of pruning in practice.

\subsection{Regularization for Pruning}
\label{sec:regularization}

Regularization in deep learning can be used for different purposes. In this work, we use $L_1$ regularization to zero out most parameters so that only a small fraction of parameters would be doing most of the recognition work. One can think of this as creating a very competitive environment where recognition resources are accumulated in a minority of kernels, and we evaluate the proposed pruning heuristics in terms of their indication quality for such a minority. 

As explained in Section \ref{sec:pruning_heuristics}, all pruning heuristics apart from weight are controlled by eigenvalues. $L_1$ regularization is a known technique to make the weights sparse \cite{goodfellow2016deep} and, naturally, make the average absolute weight small. In this section, we will prove that it can also be used to make eigenvalues small.

Let kernel $K$ be a $k \times k$ real matrix,
\begin{equation}
K = \begin{pmatrix}
    w_{11}	& w_{12}	& \dots		& w_{1n} \\
	w_{21}	& w_{22} 	& \dots		& \vdots \\
    \vdots  & \vdots	& \ddots	& \vdots \\
	w_{n1}	& \dots		& \dots		& w_{nn} \\
\end{pmatrix}
\end{equation}
define radius $r_i$ as the sum of the absolute values of the off-diagonal elements per row,
\begin{equation}
r_i = \sum_{j \neq i}^k \abs{w_{ij}}
\end{equation}
and let Gershgorin disk $D_i$ be the closed disk in the complex plane centered at $w_{ii}$ with radius $r_i$,
\begin{equation}
D_i = \{ x \in \mathbb{C}: \abs{x - w_{ii}} \leq r_i \}
\end{equation}
Then, Gershgorin Circle Theorem \cite{gershgorin1931uber} states that every eigenvalue of the kernel $K$ lies within at least one of the Gershgorin disks $D_i$.

As the weights get smaller, the origins of disks $D_i$ move towards zero. In addition, since the off-diagonal row sums get smaller, the disks shrink. In other words, excessive $L_1$ regularization collapses all Gershgorin disks $D_i$ into points near zero. 

In conclusion, $L_1$ regularization minimizes (1) the average absolute weights, and (2) the probability of having large eigenvalues. We will microscope the subtle difference between the two in our analysis sections.

\section{Experimental Results}
In this section, we present our experimental results and analyses. We first compare pruning heuristics in terms of classification accuracy and pruning ratio. In addition, we discuss real vs. complex eigenvalues for the relevant heuristics. Then, we show which heuristics prune which kernels, and how the pruning ratio changes across layers and through epochs. Each section below provides a \emph{Summary of Findings} that summarizes the corresponding experimental analysis.

\subsection{Performance Comparison}
\label{sec:performance}

In this section, we compare the pruning heuristics in terms of classification accuracy and pruning ratio which is,

\begin{equation}
    \frac{\textnormal{\# of pruned weights}}{\textnormal{total \# of weights}}
\end{equation}

At the end of a training, we first evaluate the classification accuracy without pruning ($acc_{vanilla}$), then each pruning heuristic branches off from that same parameter state to evaluate their performance. In other words, trained weights are the same for all heuristics, but the set of pruned kernels changes.

If you trim a neural net and preserver the classification accuracy, it is a good indicator that the thrown-out parts were not necessary in the first place. However, we do not search for a set of desirable hyperparameters for each pruning heuristic to get $\frac{acc_{pruned}}{acc_{vanilla}} \approx 1$. Instead, we set the environment where at least one pruning heuristic is able to prune the network while achieving competitive performance to the vanilla network.

In Table \ref{tab:performance_acc}, we see very similar performance measurements for heuristics based on spectral radius, spectral norm and the average absolute weight. The worst results obtained by the heuristics that check the smallest absolute eigenvalue (min\_eig). Since the determinant combines both the negative and the positive results through multiplication of eigenvalues, det and det\_gram heuristics oscillate between the extremums. \textbf{This finding rules out the hypothesis that an expert kernel must be a full-rank matrix under the assumption that a value smaller than the selected significance threshold is effectively $0$.} However, since the values are not \textbf{exactly} $0$, spectral radius often protects determinant from an overkill. We can clearly see the case where this is not enough in ResNet50 results where the gap between the extremums is large enough to significantly deteriorate the performance of determinant based heuristics. In Table \ref{tab:performance_ratio}, we observe the opposite, but more importantly, when we examine the ResNet50 case (especially the imagenet\_init results), it is clear that any pruning heuristic that takes the minimum absolute eigenvalue into account can prune many useful kernels and lose a significant portion of the generalization performance. Therefore, we turn our attention to \textbf{spectral radius, spectral norm and the average absolute weight}.

\begin{table*}
	 \caption{Classification accuracy comparison of different pruning heuristics}
	 \label{tab:performance_acc}
	 \centering
	 \resizebox{1\textwidth}{!}{\begin{tabular}{c|l|cc|cc|cc|ccc}

		 \toprule
		 & \multirow{2}{*}{\begin{tabular}{c}\\pruning heuristic\end{tabular}} & \multicolumn{2}{c|}{ResNet32} & \multicolumn{2}{c|}{ResNet56} & \multicolumn{2}{c|}{ResNet110} & \multicolumn{3}{c}{ResNet50} \\
		 \cmidrule{3-11}
		 & & static & random & static & random & static & random & static & random & imagenet \\
		 \midrule
		 \parbox[t]{2mm}{\multirow{8.5}{*}{\rotatebox[origin=c]{90}{CIFAR-10}}}
		 & None & $89.34 \pm 0.2$ & $89.20 \pm 0.3$ & $89.21 \pm 0.2$ & $89.19 \pm 0.2$ & $89.08 \pm 0.3$ & $88.58 \pm 0.5$ & $76.52 \pm 0.3$ & $76.32 \pm 0.4$ & $86.93 \pm 0.2$ \\
		 \cmidrule{2-11}
		 & min\_eig & $66.22 \pm 7.5$ & $65.48 \pm 9.5$ & $65.03 \pm 8.8$ & $63.84 \pm 10.9$ & $67.13 \pm 7.4$ & $65.10 \pm 10.5$ & $18.08 \pm 5.2$ & $23.70 \pm 4.9$ & $13.75 \pm 2.1$ \\
		 & min\_eig\_real & $62.83 \pm 8.1$ & $63.42 \pm 10.6$ & $62.77 \pm 9.0$ & $59.91 \pm 10.1$ & $63.81 \pm 8.5$ & $62.63 \pm 11.2$ & $17.59 \pm 5.1$ & $23.30 \pm 5.1$ & $13.53 \pm 2.0$ \\
		 & det & $89.32 \pm 0.2$ & $89.10 \pm 0.3$ & $89.13 \pm 0.2$ & $89.10 \pm 0.2$ & $89.06 \pm 0.3$ & $88.54 \pm 0.5$ & $66.17 \pm 5.9$ & $68.74 \pm 2.2$ & $56.96 \pm 4.3$ \\
		 & det\_gram & $89.31 \pm 0.2$ & $89.15 \pm 0.2$ & $89.16 \pm 0.2$ & $89.15 \pm 0.2$ & $\textbf{89.08} \pm 0.3$ & $88.54 \pm 0.5$ & $66.25 \pm 5.9$ & $68.78 \pm 2.2$ & $57.13 \pm 4.2$ \\
		 & spectral\_radius & $\textbf{89.35} \pm 0.2$ & $\textbf{89.20} \pm 0.3$ & $\textbf{89.21} \pm 0.2$ & $89.18 \pm 0.2$ & $89.07 \pm 0.3$ & $\textbf{88.58} \pm 0.5$ & $\textbf{76.34} \pm 0.6$ & $\textbf{76.12} \pm 0.7$ & $86.89 \pm 0.3$ \\
		 & spectral\_radius\_real & $89.21 \pm 0.2$ & $89.10 \pm 0.3$ & $89.07 \pm 0.2$ & $89.08 \pm 0.3$ & $88.99 \pm 0.3$ & $88.46 \pm 0.4$ & $\textbf{76.34} \pm 0.5$ & $76.06 \pm 0.6$ & $86.79 \pm 0.2$ \\
		 & spectral\_norm & $\textbf{89.35} \pm 0.2$ & $\textbf{89.20} \pm 0.3$ & $\textbf{89.21} \pm 0.2$ & $\textbf{89.19} \pm 0.2$ & $\textbf{89.08} \pm 0.3$ & $\textbf{88.58} \pm 0.5$ & $76.31 \pm 0.7$ & $\textbf{76.11} \pm 0.7$ & $\textbf{86.92} \pm 0.3$ \\
		 & weight & $\textbf{89.35} \pm 0.2$ & $\textbf{89.20} \pm 0.3$ & $\textbf{89.21} \pm 0.2$ & $89.18 \pm 0.2$ & $89.07 \pm 0.3$ & $\textbf{88.58} \pm 0.5$ & $76.19 \pm 0.7$ & $76.05 \pm 0.6$ & $\textbf{86.92} \pm 0.3$ \\
		 \midrule
		 \parbox[t]{2mm}{\multirow{8.5}{*}{\rotatebox[origin=c]{90}{CIFAR-100}}}
		 & None & $64.87 \pm 0.4$ & $64.67 \pm 0.5$ & $63.77 \pm 0.5$ & $63.89 \pm 0.4$ & $61.85 \pm 1.0$ & $61.47 \pm 0.9$ & $44.96 \pm 0.5$ & $44.70 \pm 0.5$ & $59.97 \pm 0.3$ \\
		 \cmidrule{2-11}
		 & min\_eig & $42.65 \pm 3.9$ & $38.59 \pm 5.6$ & $42.02 \pm 4.4$ & $42.84 \pm 4.1$ & $45.97 \pm 5.2$ & $46.05 \pm 3.5$ & $1.30 \pm 0.3$ & $1.31 \pm 0.4$ & $1.26 \pm 0.2$ \\
		 & min\_eig\_real & $40.25 \pm 3.6$ & $36.02 \pm 5.9$ & $39.60 \pm 5.0$ & $40.71 \pm 3.7$ & $44.19 \pm 4.5$ & $43.67 \pm 3.2$ & $1.28 \pm 0.3$ & $1.30 \pm 0.4$ & $1.27 \pm 0.2$ \\
		 & det & $64.82 \pm 0.4$ & $64.62 \pm 0.5$ & $63.62 \pm 0.6$ & $63.82 \pm 0.4$ & $61.80 \pm 1.0$ & $61.46 \pm 0.9$ & $16.32 \pm 5.4$ & $17.56 \pm 3.8$ & $9.29 \pm 1.3$ \\
		 & det\_gram & $64.83 \pm 0.4$ & $64.64 \pm 0.5$ & $63.66 \pm 0.5$ & $63.86 \pm 0.4$ & $61.79 \pm 1.0$ & $\textbf{61.47} \pm 0.9$ & $16.44 \pm 5.4$ & $17.78 \pm 3.8$ & $9.43 \pm 1.3$ \\
		 & spectral\_radius & $\textbf{64.87} \pm 0.4$ & $\textbf{64.67} \pm 0.5$ & $\textbf{63.77} \pm 0.5$ & $63.88 \pm 0.4$ & $61.84 \pm 1.0$ & $\textbf{61.48} \pm 0.9$ & $43.89 \pm 0.6$ & $43.81 \pm 0.4$ & $59.32 \pm 0.4$ \\
		 & spectral\_radius\_real & $64.74 \pm 0.4$ & $64.51 \pm 0.5$ & $63.68 \pm 0.6$ & $63.81 \pm 0.4$ & $61.79 \pm 1.0$ & $61.44 \pm 0.9$ & $43.83 \pm 0.5$ & $43.75 \pm 0.4$ & $58.56 \pm 0.4$ \\
		 & spectral\_norm & $\textbf{64.87} \pm 0.4$ & $\textbf{64.67} \pm 0.5$ & $\textbf{63.77} \pm 0.5$ & $\textbf{63.89} \pm 0.4$ & $\textbf{61.85} \pm 1.0$ & $\textbf{61.47} \pm 0.9$ & $\textcolor{blue}{43.95} \pm 0.6$ & $\textcolor{blue}{43.86} \pm 0.4$ & $\textbf{59.97} \pm 0.3$ \\
		 & weight & $\textbf{64.87} \pm 0.4$ & $64.66 \pm 0.5$ & $\textbf{63.77} \pm 0.5$ & $63.88 \pm 0.4$ & $61.84 \pm 1.0$ & $\textbf{61.47} \pm 0.9$ & $\textcolor{red}{40.19} \pm 2.4$ & $\textcolor{red}{39.40} \pm 4.1$ & $59.90 \pm 0.4$ \\
		 \midrule
		 \parbox[t]{2mm}{\multirow{8.5}{*}{\rotatebox[origin=c]{90}{tiny-imagenet}}}
		 & None & $46.62 \pm 0.4$ & $46.45 \pm 0.6$ & $45.55 \pm 0.6$ & $45.42 \pm 0.5$ & $43.75 \pm 0.9$ & $43.80 \pm 1.1$ & $42.48 \pm 1.1$ & $42.01 \pm 1.3$ & $54.41 \pm 0.4$ \\
		 \cmidrule{2-11}
		 & min\_eig & $14.40 \pm 3.6$ & $12.62 \pm 4.0$ & $15.54 \pm 3.2$ & $13.80 \pm 4.6$ & $17.81 \pm 4.4$ & $17.20 \pm 4.3$ & $26.59 \pm 1.9$ & $28.14 \pm 2.5$ & $13.45 \pm 1.1$ \\
		 & min\_eig\_real & $12.42 \pm 3.7$ & $10.73 \pm 4.1$ & $14.89 \pm 3.5$ & $12.90 \pm 4.2$ & $15.87 \pm 4.4$ & $15.85 \pm 4.7$ & $23.78 \pm 2.1$ & $25.67 \pm 2.7$ & $11.25 \pm 1.4$ \\
		 & det & $46.51 \pm 0.4$ & $46.37 \pm 0.5$ & $45.53 \pm 0.6$ & $45.26 \pm 0.5$ & $43.67 \pm 0.9$ & $43.78 \pm 1.1$ & $42.08 \pm 1.3$ & $41.66 \pm 1.3$ & $45.82 \pm 0.6$ \\
		 & det\_gram & $46.55 \pm 0.4$ & $46.39 \pm 0.6$ & $\textbf{45.56} \pm 0.6$ & $45.33 \pm 0.5$ & $43.70 \pm 0.9$ & $43.74 \pm 1.1$ & $42.05 \pm 1.3$ & $41.60 \pm 1.2$ & $46.65 \pm 0.5$ \\
		 & spectral\_radius & $\textbf{46.62} \pm 0.4$ & $46.44 \pm 0.6$ & $\textbf{45.56} \pm 0.6$ & $\textbf{45.42} \pm 0.5$ & $\textbf{43.76} \pm 0.9$ & $\textbf{43.80} \pm 1.1$ & $42.24 \pm 1.1$ & $\textbf{41.86} \pm 1.3$ & $53.93 \pm 0.3$ \\
		 & spectral\_radius\_real & $46.15 \pm 0.5$ & $46.22 \pm 0.5$ & $45.19 \pm 0.8$ & $44.78 \pm 0.9$ & $43.54 \pm 0.9$ & $43.67 \pm 1.1$ & $42.23 \pm 1.2$ & $41.74 \pm 1.3$ & $52.77 \pm 0.3$ \\
		 & spectral\_norm & $46.61 \pm 0.4$ & $\textbf{46.45} \pm 0.6$ & $\textbf{45.55} \pm 0.6$ & $\textbf{45.42} \pm 0.5$ & $\textbf{43.75} \pm 0.9$ & $\textbf{43.81} \pm 1.1$ & $\textbf{42.28} \pm 1.1$ & $\textbf{41.85} \pm 1.3$ & $\textbf{54.42} \pm 0.4$ \\
		 & weight & $\textbf{46.62} \pm 0.4$ & $46.44 \pm 0.6$ & $\textbf{45.56} \pm 0.6$ & $\textbf{45.42} \pm 0.5$ & $\textbf{43.75} \pm 0.9$ & $\textbf{43.81} \pm 1.1$ & $42.25 \pm 1.1$ & $\textbf{41.86} \pm 1.3$ & $\textbf{54.41} \pm 0.4$ \\
		  \bottomrule
		  \end{tabular}}
	 \vspace{-3pt}
\end{table*}

\begin{table*}
	 \caption{Pruning ratio comparison of different pruning heuristics}
	 \label{tab:performance_ratio}
	 \centering
	 \resizebox{1\textwidth}{!}{\begin{tabular}{c|l|cc|cc|cc|ccc}

		 \toprule
		 & \multirow{2}{*}{\begin{tabular}{c}\\pruning heuristic\end{tabular}} & \multicolumn{2}{c|}{ResNet32} & \multicolumn{2}{c|}{ResNet56} & \multicolumn{2}{c|}{ResNet110} & \multicolumn{3}{c}{ResNet50} \\
		 \cmidrule{3-11}
		 & & static & random & static & random & static & random & static & random & imagenet \\
		 \midrule
		 \parbox[t]{2mm}{\multirow{8.5}{*}{\rotatebox[origin=c]{90}{CIFAR-10}}}
		 & min\_eig & $74.54 \pm 0.4$ & $75.36 \pm 0.8$ & $85.67 \pm 0.3$ & $85.81 \pm 0.4$ & $92.17 \pm 0.2$ & $92.25 \pm 0.3$ & $93.06 \pm 0.1$ & $93.03 \pm 0.2$ & $94.01 \pm 0.6$ \\
		 & min\_eig\_real & $\textbf{75.35} \pm 0.3$ & $\textbf{76.11} \pm 0.7$ & $\textbf{86.12} \pm 0.3$ & $\textbf{86.26} \pm 0.4$ & $\textbf{92.39} \pm 0.2$ & $\textbf{92.47} \pm 0.3$ & $\textbf{93.12} \pm 0.1$ & $\textbf{93.09} \pm 0.2$ & $\textbf{94.12} \pm 0.6$ \\
		 & det & $69.69 \pm 0.5$ & $70.63 \pm 1.0$ & $82.90 \pm 0.4$ & $83.02 \pm 0.6$ & $90.84 \pm 0.3$ & $90.87 \pm 0.4$ & $92.62 \pm 0.1$ & $92.57 \pm 0.2$ & $93.22 \pm 0.6$ \\
		 & det\_gram & $69.67 \pm 0.5$ & $70.62 \pm 1.0$ & $82.90 \pm 0.4$ & $83.01 \pm 0.6$ & $90.84 \pm 0.3$ & $90.86 \pm 0.4$ & $92.62 \pm 0.1$ & $92.57 \pm 0.2$ & $93.21 \pm 0.6$ \\
		 & spectral\_radius & $67.88 \pm 0.6$ & $68.83 \pm 1.2$ & $81.82 \pm 0.5$ & $81.96 \pm 0.6$ & $90.31 \pm 0.3$ & $90.33 \pm 0.5$ & $90.96 \pm 0.4$ & $90.77 \pm 0.4$ & $89.53 \pm 1.0$ \\
		 & spectral\_radius\_real & $68.30 \pm 0.6$ & $69.24 \pm 1.1$ & $82.08 \pm 0.5$ & $82.20 \pm 0.6$ & $90.43 \pm 0.3$ & $90.45 \pm 0.5$ & $91.06 \pm 0.4$ & $90.87 \pm 0.4$ & $89.72 \pm 1.0$ \\
		 & spectral\_norm & $67.22 \pm 0.6$ & $68.16 \pm 1.2$ & $81.41 \pm 0.5$ & $81.59 \pm 0.7$ & $90.12 \pm 0.3$ & $90.15 \pm 0.5$ & $90.72 \pm 0.4$ & $90.50 \pm 0.4$ & $88.90 \pm 1.0$ \\
		 & weight & $68.32 \pm 0.6$ & $69.24 \pm 1.1$ & $82.09 \pm 0.4$ & $82.22 \pm 0.6$ & $90.45 \pm 0.3$ & $90.48 \pm 0.5$ & $91.68 \pm 0.3$ & $91.52 \pm 0.3$ & $90.91 \pm 0.8$ \\
		 \midrule
		 \parbox[t]{2mm}{\multirow{8.5}{*}{\rotatebox[origin=c]{90}{CIFAR-100}}}
		 & min\_eig & $44.89 \pm 1.1$ & $48.53 \pm 2.1$ & $74.66 \pm 0.7$ & $74.37 \pm 1.4$ & $87.96 \pm 1.0$ & $88.01 \pm 0.7$ & $81.50 \pm 0.9$ & $80.76 \pm 1.3$ & $92.68 \pm 0.2$ \\
		 & min\_eig\_real & $\textbf{46.41} \pm 1.1$ & $\textbf{49.87} \pm 2.0$ & $\textbf{75.25} \pm 0.6$ & $\textbf{74.94} \pm 1.4$ & $\textbf{88.20} \pm 0.9$ & $\textbf{88.23} \pm 0.6$ & $\textbf{81.64} \pm 0.9$ & $\textbf{80.91} \pm 1.3$ & $\textbf{92.75} \pm 0.2$ \\
		 & det & $35.18 \pm 1.4$ & $39.93 \pm 2.8$ & $71.09 \pm 0.9$ & $70.84 \pm 1.8$ & $86.57 \pm 1.2$ & $86.66 \pm 0.9$ & $78.25 \pm 0.9$ & $77.36 \pm 1.3$ & $91.95 \pm 0.2$ \\
		 & det\_gram & $35.15 \pm 1.4$ & $39.90 \pm 2.8$ & $71.07 \pm 0.9$ & $70.83 \pm 1.8$ & $86.56 \pm 1.3$ & $86.65 \pm 0.9$ & $78.24 \pm 0.9$ & $77.35 \pm 1.3$ & $91.94 \pm 0.2$ \\
		 & spectral\_radius & $33.01 \pm 1.6$ & $38.12 \pm 3.0$ & $70.29 \pm 1.0$ & $70.11 \pm 1.9$ & $86.19 \pm 1.3$ & $86.29 \pm 0.9$ & $67.50 \pm 2.2$ & $66.25 \pm 2.6$ & $82.55 \pm 0.4$ \\
		 & spectral\_radius\_real & $33.60 \pm 1.5$ & $38.61 \pm 3.0$ & $70.51 \pm 1.0$ & $70.31 \pm 1.9$ & $86.28 \pm 1.3$ & $86.37 \pm 0.9$ & $67.78 \pm 2.2$ & $66.53 \pm 2.6$ & $82.72 \pm 0.4$ \\
		 & spectral\_norm & $32.44 \pm 1.6$ & $37.64 \pm 3.2$ & $70.05 \pm 1.0$ & $69.89 \pm 1.9$ & $86.07 \pm 1.3$ & $86.16 \pm 0.9$ & $67.03 \pm 2.2$ & $65.76 \pm 2.6$ & $81.38 \pm 0.4$ \\
		 & weight & $33.46 \pm 1.5$ & $38.48 \pm 3.0$ & $70.45 \pm 1.0$ & $70.26 \pm 1.9$ & $86.28 \pm 1.3$ & $86.37 \pm 0.9$ & $73.19 \pm 1.2$ & $72.03 \pm 1.7$ & $83.92 \pm 0.4$ \\
		 \midrule
		 \parbox[t]{2mm}{\multirow{8.5}{*}{\rotatebox[origin=c]{90}{tiny-imagenet}}}
		 & min\_eig & $68.83 \pm 0.5$ & $68.73 \pm 1.0$ & $83.35 \pm 0.6$ & $83.57 \pm 0.5$ & $91.63 \pm 0.4$ & $91.42 \pm 0.7$ & $84.30 \pm 1.3$ & $83.19 \pm 1.1$ & $91.85 \pm 0.0$ \\
		 & min\_eig\_real & $\textbf{69.46} \pm 0.5$ & $\textbf{69.43} \pm 1.0$ & $\textbf{83.63} \pm 0.6$ & $\textbf{83.85} \pm 0.4$ & $\textbf{91.74} \pm 0.4$ & $\textbf{91.55} \pm 0.6$ & $\textbf{84.95} \pm 1.1$ & $\textbf{83.90} \pm 1.0$ & $\textbf{92.00} \pm 0.0$ \\
		 & det & $64.22 \pm 0.8$ & $63.77 \pm 1.5$ & $81.35 \pm 0.8$ & $81.56 \pm 0.7$ & $90.87 \pm 0.5$ & $90.55 \pm 0.9$ & $81.72 \pm 1.7$ & $80.56 \pm 1.6$ & $90.76 \pm 0.0$ \\
		 & det\_gram & $64.20 \pm 0.8$ & $63.76 \pm 1.5$ & $81.35 \pm 0.8$ & $81.55 \pm 0.7$ & $90.87 \pm 0.5$ & $90.55 \pm 0.9$ & $81.72 \pm 1.7$ & $80.56 \pm 1.6$ & $90.74 \pm 0.0$ \\
		 & spectral\_radius & $63.52 \pm 0.8$ & $62.95 \pm 1.6$ & $81.02 \pm 0.8$ & $81.22 \pm 0.7$ & $90.73 \pm 0.5$ & $90.40 \pm 1.0$ & $80.30 \pm 2.1$ & $79.00 \pm 2.1$ & $86.71 \pm 0.0$ \\
		 & spectral\_radius\_real & $63.81 \pm 0.8$ & $63.28 \pm 1.6$ & $81.15 \pm 0.8$ & $81.34 \pm 0.7$ & $90.77 \pm 0.5$ & $90.45 \pm 0.9$ & $80.64 \pm 2.0$ & $79.37 \pm 2.0$ & $86.98 \pm 0.0$ \\
		 & spectral\_norm & $63.36 \pm 0.8$ & $62.76 \pm 1.6$ & $80.94 \pm 0.8$ & $81.13 \pm 0.8$ & $90.68 \pm 0.5$ & $90.35 \pm 1.0$ & $79.88 \pm 2.2$ & $78.54 \pm 2.3$ & $84.97 \pm 0.1$ \\
		 & weight & $63.64 \pm 0.8$ & $63.10 \pm 1.6$ & $81.09 \pm 0.8$ & $81.28 \pm 0.7$ & $90.76 \pm 0.5$ & $90.43 \pm 1.0$ & $81.28 \pm 1.5$ & $80.12 \pm 1.5$ & $86.78 \pm 0.0$ \\
		  \bottomrule
		  \end{tabular}}
	 \vspace{-3pt}
\end{table*}

Summary of findings:
\begin{itemize}
    \item Any heuristic that uses the min abs eigenvalue is bad.
    
    \item Expert kernel $\neq$ full-rank matrix.
    
    \item Spectral norm is the strongest candidate in terms of preserving the vanilla classification accuracy since it prunes the smallest subset of kernels in comparison (see Section \ref{sec:set_analysis}).
\end{itemize}

\subsection{Real vs. Complex Eigenvalues}
\label{sec:eig_analysis}

Here, we are interested in the relevance of the real and complex parts of eigenvalues to detect expert kernels. Therefore, we have min\_eig\_real and spectral\_radius\_real heuristics in addition to their counterparts which that take the complex parts into account as well. The real-part-only heuristics are expected to prune more kernels than their vanilla counterparts since
\begin{equation}
    \sqrt{a^2 + b^2} \geq a,
\end{equation}
where the complex eigenvalue $\lambda = a + bi$. Moreover, when we only consider the real parts of the eigenvalues, their ordering may change which means, for instance, min\_eig may target $\lambda_1$ whereas min\_eig\_real targets $\lambda_2$ for the same kernel, where $\abs{\lambda_1} < \abs{\lambda_2}$ but $\abs{a_1} > \abs{a_2}$.

Naturally, to be able to test the respective relevance of real and complex parts, we need to have a significant amount of complex eigenvalues in our CNNs. In Table \ref{tab:eig_stats_thin}, we show the ratio of complex eigenvalues from three angles: (1) how many complex eigenvalues values do we have? ($\frac{\textnormal{\# of complex eigvals}}{\textnormal{\# of total eigvals}}$), (2) how many complex eigenvalues are \textbf{targeted} by the relevant pruning heuristics? ($\frac{\textnormal{\# of targeted complex eigvals}}{\textnormal{\# of targeted total eigvals}}$), and (3) how many \textbf{pruned} kernels are deemed unimportant by looking at complex eigenvalues? ($\frac{\textnormal{\# of pruned kernels using complex eigvals}}{\textnormal{\# of pruned kernels}}$).

For ResNet50, $1 \times 1$ kernels dominates the picture, and, since they do not have any complex eigenvalues, ResNet50 does not contain a significant amount of complex eigenvalues for an analysis ($\approx2 \%$ of pruned eigenvalues).

For thin ResNets, total amount of complex eigenvalues range from $30\%$ to $40\%$ consistently across different models, which is enough to affect the overall classification performance (Table \ref{tab:eig_stats_thin}). However, the total complex/real ratio may not be quite informative since different eigenvalue-based pruning heuristics focus on different eigenvalues. For example, if the smallest absolute eigenvalues of all kernels are real and all the largest ones are complex, min\_eig and spectral\_radius may cause significantly different behavior in terms of vanilla heuristics vs. real value based counterparts. Hence, we include a category called \textbf{target} which indicates the amount of complex values evaluated by each pruning heuristic whether they lead to pruning or not. According to Table \ref{tab:eig_stats_thin}, there is indeed a significant difference in the sets of targeted values by vanilla and real value based heuristics, and this observation is true for both min\_eig and spectral\_radius. Thus, when we check the results in Table \ref{tab:performance_acc}, we see that min\_eig\_real has a noticeable disadvantage against min\_eig. Nevertheless, spectral\_radius remains stable whether we choose to focus only on real values or not. This, of course, holds no importance if we did not prune any kernel due to its complex eigenvalue in the first place. Therefore, we also show the overall contribution of complex eigenvalues to the pruning. The fact that $20\% - 40\%$ of the pruned kernels are selected by looking at complex values demonstrates that there were indeed enough room for a significant change in pruning performance between vanilla and real-part-only heuristics (Section \ref{sec:performance}).

At the end, we see that ignoring the complex parts leads to worse performance for min\_eig heuristic whereas it has no significant effect on the performance of spectral\_radius heuristic. However, since min\_eig has already started to prune essential kernels, it is no surprise that min\_eig\_real prunes even more essential kernels. This is not the case for spectral\_radius. For example, in Table \ref{tab:performance_ratio}, a slight increase ($0.22 \%$) in pruning ratio due to min\_eig\_real of ResNet110 (with static initialization) causes a $3.32 \%$ drop in CIFAR-10 classification accuracy. However, for the same case, spectral\_radius\_real provides a $0.12 \%$ increase in pruning ratio for only $0.08 \%$ drop in accuracy. We think that this may prove useful for practitioners who want to push the pruning ratio as much as possible while preserving most of the recognition performance.

\begin{table*}
	 \caption{Amount of total/targeted/pruned complex eigenvalues}
	 \label{tab:eig_stats_thin}
	 \centering
	 \resizebox{1\textwidth}{!}{\begin{tabular}{c|c|c|cc|cc|cc|cc}

		 \toprule
		 & \multirow{2}{*}{\begin{tabular}{c}\\model\end{tabular}} & \multirow{2}{*}{\begin{tabular}{c}\vspace{-9pt}\\total complex\\ ratio\end{tabular}} & \multicolumn{2}{c|}{min\_eig} & \multicolumn{2}{c|}{min\_eig\_real} & \multicolumn{2}{c|}{spectral\_radius} & \multicolumn{2}{c}{spectral\_radius\_real} \\
		 \cmidrule{4-11}
		 & & & target & pruned & target & pruned & target & pruned & target & pruned \\
		 \midrule
		 \parbox[t]{2mm}{\multirow{3}{*}{\rotatebox[origin=c]{90}{\small{C-10}}}}
		 & ResNet32 & $36.3 \pm 0.2$  & $24.7 \pm 0.2$ & $26.5 \pm 0.3$ & $36.0 \pm 0.3$ & $39.3 \pm 0.3$ & $28.1 \pm 0.3$ & $32.9 \pm 0.3$ & $16.8 \pm 0.2$ & $20.0 \pm 0.2$ \\
		 & ResNet56 & $39.1 \pm 0.1$  & $26.8 \pm 0.1$ & $28.0 \pm 0.2$ & $39.1 \pm 0.2$ & $41.2 \pm 0.2$ & $30.9 \pm 0.1$ & $33.8 \pm 0.1$ & $18.6 \pm 0.1$ & $20.5 \pm 0.1$ \\
		 & ResNet110 & $41.0 \pm 0.1$  & $28.2 \pm 0.1$ & $28.9 \pm 0.1$ & $41.2 \pm 0.2$ & $42.4 \pm 0.2$ & $32.8 \pm 0.1$ & $34.3 \pm 0.1$ & $19.7 \pm 0.1$ & $20.6 \pm 0.1$ \\
		 \midrule
		 \parbox[t]{2mm}{\multirow{3}{*}{\rotatebox[origin=c]{90}{\small{C-100}}}}
		 & ResNet32 & $31.3 \pm 0.2$  & $21.9 \pm 0.1$ & $24.5 \pm 0.2$ & $31.2 \pm 0.1$ & $37.4 \pm 0.3$ & $23.6 \pm 0.4$ & $34.3 \pm 0.5$ & $14.3 \pm 0.2$ & $21.2 \pm 0.1$ \\
		 & ResNet56 & $36.6 \pm 0.2$  & $25.3 \pm 0.1$ & $27.6 \pm 0.1$ & $36.8 \pm 0.2$ & $41.0 \pm 0.2$ & $28.6 \pm 0.2$ & $34.2 \pm 0.1$ & $17.1 \pm 0.1$ & $20.6 \pm 0.1$ \\
		 & ResNet110 & $40.5 \pm 0.1$  & $27.9 \pm 0.1$ & $28.9 \pm 0.1$ & $40.8 \pm 0.2$ & $42.6 \pm 0.1$ & $32.3 \pm 0.1$ & $34.4 \pm 0.1$ & $19.4 \pm 0.1$ & $20.6 \pm 0.0$ \\
		 \midrule
		 \parbox[t]{2mm}{\multirow{3}{*}{\rotatebox[origin=c]{90}{\small{t-img}}}}
		 & ResNet32 & $35.5 \pm 0.3$  & $24.0 \pm 0.2$ & $26.1 \pm 0.4$ & $35.2 \pm 0.3$ & $39.0 \pm 0.4$ & $27.5 \pm 0.3$ & $32.4 \pm 0.2$ & $16.3 \pm 0.1$ & $19.4 \pm 0.2$ \\
		 & ResNet56 & $39.1 \pm 0.2$  & $26.7 \pm 0.4$ & $28.0 \pm 0.3$ & $39.4 \pm 0.3$ & $41.5 \pm 0.2$ & $30.8 \pm 0.1$ & $33.2 \pm 0.3$ & $18.2 \pm 0.1$ & $19.6 \pm 0.1$ \\
		 & ResNet110 & $40.7 \pm 0.1$  & $28.0 \pm 0.1$ & $28.5 \pm 0.1$ & $41.3 \pm 0.1$ & $42.2 \pm 0.1$ & $32.5 \pm 0.1$ & $33.5 \pm 0.1$ & $19.2 \pm 0.1$ & $19.8 \pm 0.1$ \\
		 \bottomrule
		 \end{tabular}}
\end{table*}

Summary of findings:
\begin{itemize}
    \item We have not found a major difference between using spectral\_radius and spectral\_radius\_real in terms of preserved classification accuracy. Hence, in some cases, spectral\_radius\_real may be used to prune more kernels while retaining most of the generalization performance.
    
    \item Since min\_eig has already started to prune expert kernels, min\_eig\_real makes the performance even worse by pruning even more of them.
\end{itemize}

\subsection{Set Analysis}
\label{sec:set_analysis}

In this section, we examine the sets of kernels pruned by each pruning heuristic and their intersections. In our experiments, for almost all datasets, models, and initializations, there are only 10 distinct sets of kernels which are pruned by:
\begin{enumerate}
    \item (min\_eig, weight, det, spectral\_radius, spectral\_norm, det\_gram)
    \item (min\_eig, weight, det, spectral\_radius, det\_gram)
    \item (min\_eig, weight, det, det\_gram)
    \item (min\_eig, det, spectral\_radius, det\_gram)
    \item (min\_eig, det, det\_gram)
    \item (min\_eig, det\_gram)
\end{enumerate}
\vspace{-6mm}
\begin{multicols}{2}
    \begin{enumerate}
        \setcounter{enumi}{6}
        \item (min\_eig, weight)
        \item (min\_eig, det)
        \item min\_eig
        \item weight
    \end{enumerate}
\end{multicols}

Except 5 cases: (1,2) ResNet50 and ResNet56 for CIFAR-100 using static\_init, (3) ResNet32 for tiny-imagenet using random\_init, and (4) ResNet50 for tiny-imagenet using imagenet\_init. These cases also have kernels pruned by (min\_eig, det, spectral\_radius). In addition, (5) we may also observe kernels pruned only by det\_gram due to numerical instability born out of dealing with very small numbers.

We can draw several conclusions from these sets $S$:
\begin{itemize}
    \item The kernels pruned by spectral norm forms the subset of all other sets, so it is the safest heuristic among others:
    \begin{equation}
        S_{\textnormal{min\_eig}} \cap S_{\textnormal{weight}} \cap S_{\textnormal{det}} \cap S_{\textnormal{spectral\_radius}} \cap S_{\textnormal{det\_gram}} = S_{\textnormal{spectral\_norm}}
    \end{equation}
    
    \item The extremums created by the heuristics based on the largest and smallest absolute eigenvalues, reflect themselves as:
    \begin{equation}
        S_{\textnormal{min\_eig}} \supset S_{\textnormal{det}} \supset S_{\textnormal{spectral\_radius}}
    \end{equation}
    
    \item The numerical instability mentioned in Section \ref{sec:performance} prevents $S_{\textnormal{det}}$ and $S_{\textnormal{det\_gram}}$ to be identical. Moreover, due to the existence of (min\_eig, det) and (min\_eig, det\_gram), one is not a subset of the other either:
    \begin{equation}
    \begin{gathered}
        S_{\textnormal{det}} \neq S_{\textnormal{det\_gram}}        \\
        S_{\textnormal{det}} \not\supset S_{\textnormal{det\_gram}} \\
        S_{\textnormal{det\_gram}} \not\supset S_{\textnormal{det}}
    \end{gathered}
    \end{equation}
    Another outcome of this numerical problem:
    \begin{equation}
        S_{\textnormal{det\_gram}} \not\supset S_{\textnormal{spectral\_radius}}
    \end{equation}
    
    \item We can see the distinction between weight-based pruning and eigenvalue-based pruning:
    \begin{equation}
    \begin{gathered}
        S_{\textnormal{min\_eig}} \not\supset S_{\textnormal{weight}}        \\
        S_{\textnormal{weight}} \not\supset \{S_{\textnormal{min\_eig}}, S_{\textnormal{det}}, S_{\textnormal{spectral\_radius}}, S_{\textnormal{det\_gram}}\}
    \end{gathered}
    \end{equation}
\end{itemize}

\subsection{Pruning per Layer}
\label{sec:pruning_per_layer}

\begin{figure*}
\centering
\includegraphics[scale=0.17]{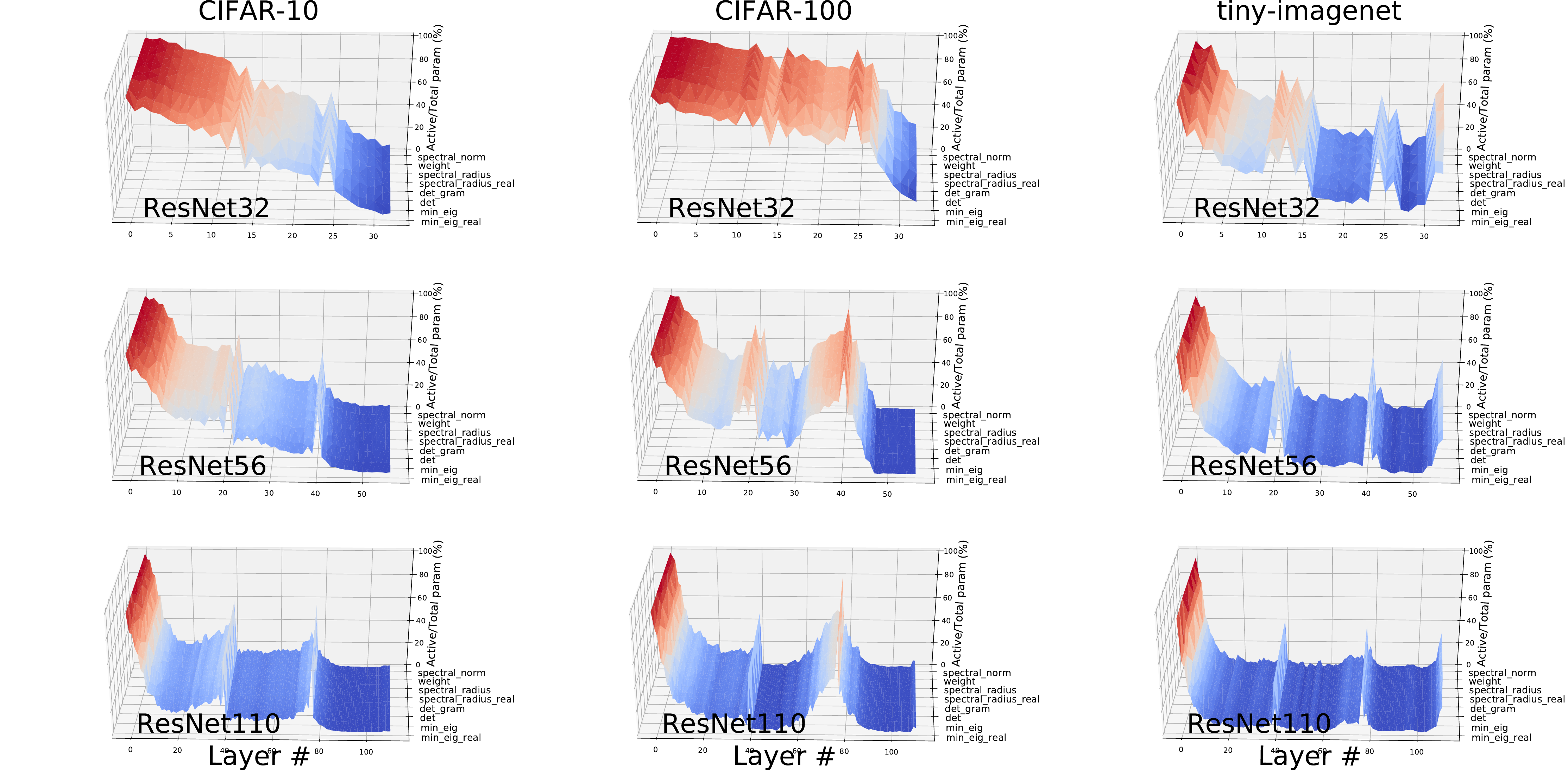}
\caption{Active parameter ratio of different pruning heuristics through layers (input $\rightarrow$ output) for thin ResNets under multiple random initializations.}
\label{fig:ThinMicroResNet_pruning_per_layer}
\end{figure*}

\begin{figure*}
\centering
\includegraphics[scale=0.17]{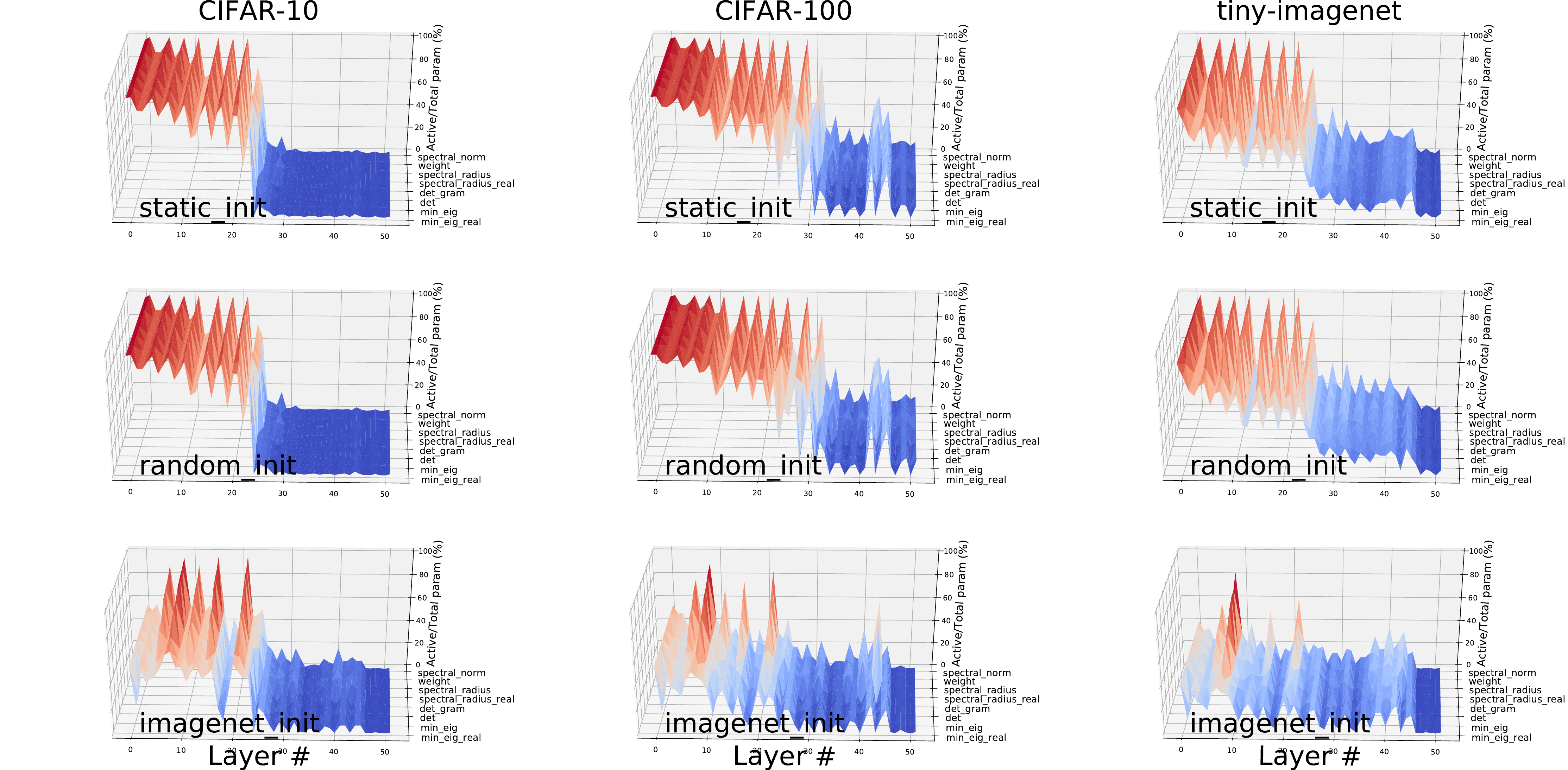}
\caption{Active parameter ratio of different pruning heuristics through layers (input $\rightarrow$ output) for ResNet50 under different initializations.}
\label{fig:MicroResNet50_pruning_per_layer}
\end{figure*}

In this section, we study how pruning affects different layers. In Fig \ref{fig:ThinMicroResNet_pruning_per_layer}, we show the amount of activity ($\frac{\textnormal{\# of active params}}{\textnormal{total \# of params}}$) in different layers of different models of thin ResNet family. Considering the similarity of overall activity pattern, we see that \textbf{the dataset at hand has a footprint over the pruned architecture}. In detail, row-wise similarity is relatively low (same model - different dataset) whereas column-wise similarity indicates a correlation of activity patterns of different models from the same family for the same dataset. Of course, as the model gets smaller (ResNet32) activity of the first layers gets higher than that of larger models (ResNet56 and ResNet110). Nevertheless, it is interesting to see \textbf{aligned spikes in activity}, for example the spike at layers: ResNet32:25 - ResNet56:40 - ResNet110:80 on CIFAR-100, since these alignments require the stretch of activies over layers for different models where the number of layers are significantly different. However, generalization of this behavior is restricted within a somewhat vague model family definition (thin ResNets) as ResNet50 does not exhibit the same overall activity pattern. It seems using fundamental building blocks (thin ResNet block) to generate shallow/deep models leads to interesting activity distributions which may be used to study the importance/effect of depth in neural nets from a different perspective. Another factor that affects the activity pattern is initialization. We show this effect in Fig \ref{fig:MicroResNet50_pruning_per_layer}. CNN training contains different sources of randomness that can affect the results. Hence, we include static\_init results to compare with random\_init results to show how little these perturbations are in our experiments. Nevertheless, the key observation here is that \textbf{significantly different initializations} (imagenet\_init vs. random\_init) \textbf{results in different activity patterns}. In addition, imagenet\_init is the only case where we observe a drop in first layers' activities. On the other hand, with the exception of sudden increase in last layer activities on tiny\_imagenet, we see a general proclivity towards low activity in last layers, but the exact reason for this phenomenon and its generality remains an open question.

Summary of findings:
\begin{itemize}
    \item Dataset has a footprint over the activity map ($\frac{\textnormal{\# of active params}}{\textnormal{total \# of params}}$) of a CNN.
    
    \item Thin ResNet experiments show that, within the same model family, activity maps of different models resemble each other.
    
    \item ResNet50 experiments show that significantly different initializations affect the activity map.
    
    \item For randomly initialized models, first layers have relatively smaller pruning ratio than the deeper layers.
\end{itemize}

\subsection{Pruning through Epochs}
\label{sec:pruning_through_epochs}

Here, we decided on the number of epochs (200) by looking at the learning curves of multiple experiments. Our policy was to keep training for a while after the learning curve became nearly flat. However, recently, the training history gained more importance due to interesting observations such as \textit{epoch-wise double descent} \cite{nakkiran2019deep}. Hence, after every $10$ epochs, we \textbf{simulated} pruning for each pruning heuristic as if it was the end of the training to construct a \textbf{pruning history}. Using this history, we ask: What would we see if we stopped the training earlier?

\begin{figure*}
\centering
\includegraphics[scale=0.32]{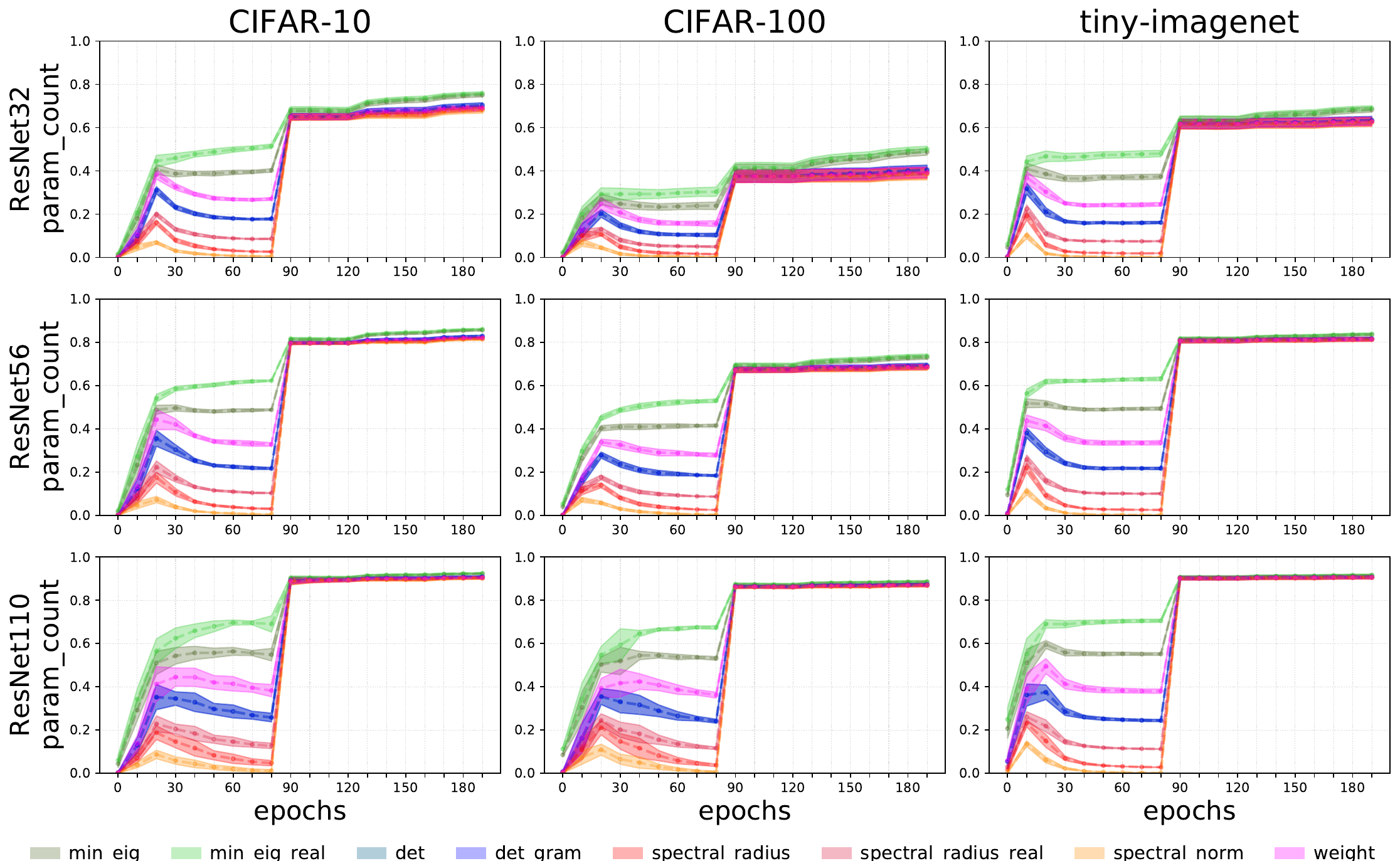}
\caption{Pruning ratios of different thin ResNets through epochs for different datasets.}
\label{fig:ThinMicroResNet_param_count_history_full}
\end{figure*} 

\begin{figure*}
\centering
\includegraphics[scale=0.32]{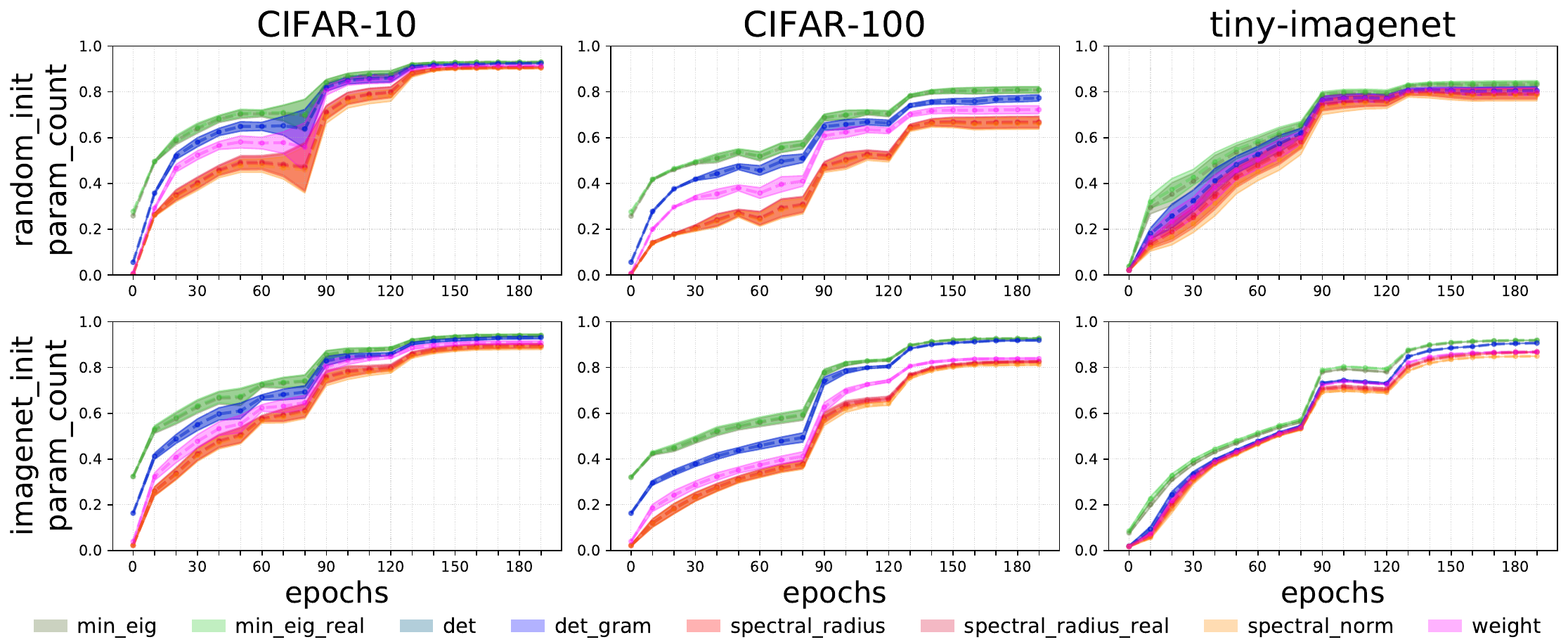}
\caption{Pruning ratios of ResNet50 through epochs for different datasets under different initializations.}
\label{fig:MicroResNet50_param_count_history_full}
\end{figure*} 

In Fig \ref{fig:ThinMicroResNet_param_count_history_full} and Fig \ref{fig:MicroResNet50_param_count_history_full}, we show pruning ratios of each pruning heuristic through epochs. As expected (see Sections \ref{sec:performance}, \ref{sec:set_analysis}, and \ref{sec:eig_analysis}), min\_eig\_real always prunes the most. The interesting observation here is that, for thin ResNets, other pruning heuristics almost catches up min\_eig in terms of pruning ratio after 90$^{\textnormal{th}}$ epoch which is right after we divide the learning rate by $10$ (80$^{\textnormal{th}}$ epoch). This is not the case for ResNet50 where the curves are quite similar to each other and the gap between them is never as big as in the thin ResNet case. However, ResNet50 contains a significant amount of $1 \times 1$ kernels (no difference between pruning heuristics), so it is expected for different heuristics to draw similar curves.

\begin{figure*}
\centering
\includegraphics[scale=0.32]{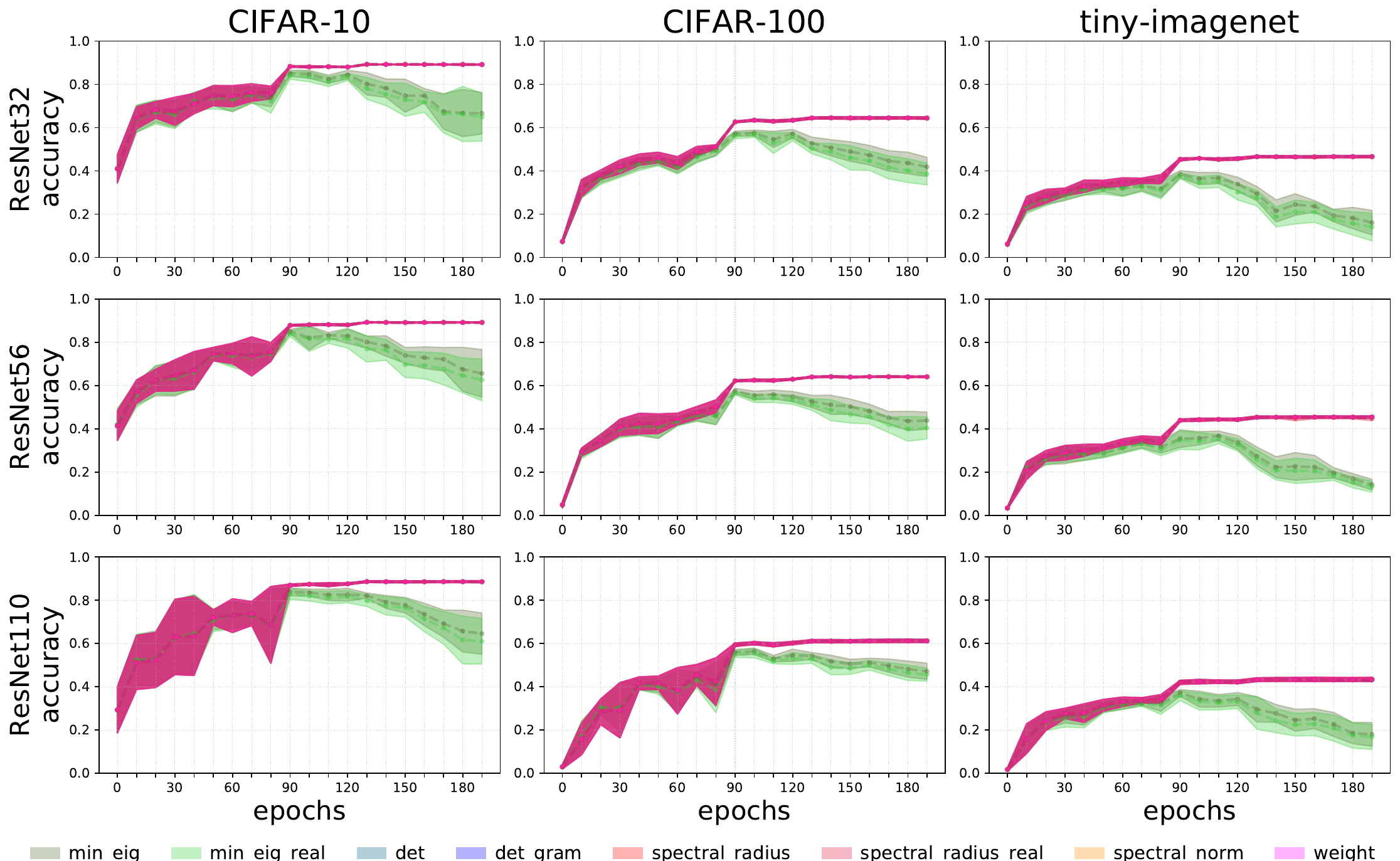}
\caption{Classification accuracy of different thin ResNets through epochs for different datasets.}
\label{fig:ThinMicroResNet_accuracy_history_full}
\end{figure*} 

\begin{figure*}
\centering
\includegraphics[scale=0.32]{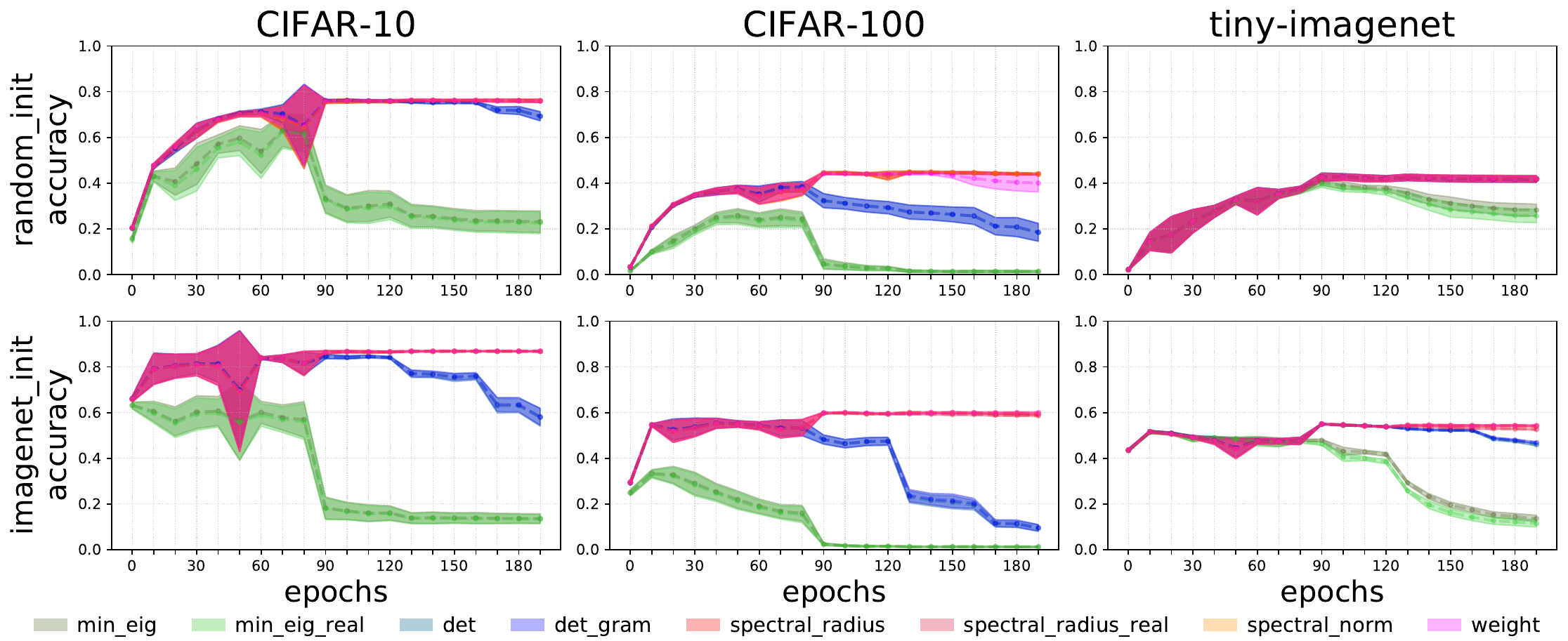}
\caption{Classification accuracy of ResNet50 through epochs for different datasets under different initializations.}
\label{fig:MicroResNet50_accuracy_history_full}
\end{figure*}

In Fig \ref{fig:ThinMicroResNet_accuracy_history_full}, we show classification accuracy of each pruning heuristic through epochs for different thin ResNets. In comparison to pruning ratio charts, classification performances vary significantly which results in indistinguishable results for the first 90 epochs. However, after that point, there is a strong divergence in performance for min\_eig and min\_eig\_real whereas other heuristics continue to perform quite similarly. Note that det and det\_gram can also diverge from the red-ish curve depending on the eigenvalue distribution, and we observe this probability in Fig \ref{fig:MicroResNet50_accuracy_history_full} as the blue-ish curves also diverge from the red-ish ones in the later epochs of training.

In overall, we found an interesting point of behavioral change in pruning, epoch $80$. \textbf{After the first drop in learning rate, we identified an interesting shift in the eigenvalue distribution resulting in (1) convergence of different heuristics' pruning scores, then (2) divergence of minimum absolute eigenvalue based heuristics}. Why is this important? Nakkiran et al. \cite{nakkiran2019deep} note that if one has a large enough model they can observe an epoch-wise double descent where the training is split into two halves with different behaviors. This \textbf{two stage learning} model also resembles the information bottleneck principle \cite{shwartz2017opening} and the critical learning period \cite{achille2018critical}. It seems that, in deep learning literature, the two phased learning is observed again and again in different concepts. Although, we cannot explain the behavior of the eigenvalues in our experiments, it is important to report this incident since each new observation of the \textbf{two stage learning} model decreases the odds of experiencing a mere coincidence. 

Summary of findings:
\begin{itemize}
    \item The first drop in learning rate is found to be a very interesting turning point for pruning:
    \begin{itemize}
        \item The difference between the min and the max abs eigenvalues of $3 \times 3$ kernels starts to increase. As a result, the heuristics that use the former starts to prune more and suffer in classification performance.
        
        \item The pruning ratios of all heuristics starts to increase until they reach approximately the same amount of pruning. Nevertheless, as the training continues the min abs eigenvalue based heuristics eventually reach higher pruning ratios.
    \end{itemize}
\end{itemize}

\section{Discussion}

We can also find expert kernels in human brain. It is well known that adult brain contains special modules that activate for specific functions such as face recognition \cite{kanwisher1997fusiform}. What is more interesting for us is that these experts emerge from a natural pruning phase \cite{national2000people}: ``In normal development, the pathway for each eye is sculpted (or “pruned”) down to the right number of connections, and those connections are sculpted in other ways, for example, to allow one to see patterns. By overproducing synapses then selecting the right connections, the brain develops an organized wiring diagram that functions optimally." A clear example is given by Haier et al. \cite{haier1992regional,haier1992intelligence} where it is shown that high brain activity while learning to play Tetris is replaced by a low brain activity once the subjects reach a certain level of expertise in the game. The authors of these papers state that the correlation between improvement on the task and decreasing brain glucose use suggests that those who honed their cognitive strategy to the fewest circuits improved the most. In fact, these studies are part of a wider picture known as the \textbf{neural efficiency hypothesis of intelligence} \cite{neubauer2009intelligence,nussbaumer2015neural}. One can draw parallels between these findings of neural efficiency and the learning dynamics in convolutional neural networks. We believe a complete  understanding of the learning dynamics of CNNs would not be complete without pruning.

\section{Conclusions and Future Work}

In this work, we study matrix characteristics of kernels for pruning convolutional neural networks in an attempt to answer the question: how can we characterize or identify expert kernels -- the kernels, which degrades generalization when removed? In this pursuit, based on a comprehensive set of experiments, we found that 

\begin{enumerate}
    \item Spectral norm is a more robust heuristic (in the sense that it better preserves generalization accuracy) than the average absolute weight for pruning. 
    \item A kernel can be expert whether it is full-rank or not. 
    \item Theoretical pruning heuristics may not hold in practice due to numerical instabilities born out of dealing with very small numbers. 
    \item Datasets have a significant footprint in the activity map (active parameter distribution across layers after pruning) of a model family in a homogeneous $L_1$ regularization (each layer has the same regularization penalty). 
    \item The smallest absolute eigenvalue of a kernel cannot be used to mark expert kernels. We provide an observation to show this where the gap between the smallest and the largest starts to increase after the first drop in the learning rate which results in a divergence of pruning performances achieved by minimum and maximum absolute eigenvalue based pruning heuristics.
\end{enumerate} 

With the above findings in mind, we leave an extensive study of the distribution of expert kernels over layers for different problems (e.g. to test the feasibility of  multi-task learning within a single model), two distinct patterns of eigenvalue distributions and their connection to the learning rate for future research. Deriving a practical pruning recipe out of these findings is also worth exploring in the future.

%\section*{References}

\bibliography{mybibfile}

\begin{frontmatter}

\title{\\Supplementary Material}

\end{frontmatter}

\section{Experimental Setup}
\label{sec:exp_setup}

\subsection{Dataset}

We used 3 generic object classification datasets: \{CIFAR-10 \cite{krizhevsky2009learning}, CIFAR-100 \cite{krizhevsky2009learning}, tiny-imagenet \cite{tinyimagenet}\}. CIFAR-10 and CIFAR-100 datasets are well known benchmarks for generic image classification. tiny-imagenet is a subset of ImageNet \cite{russakovsky2015imagenet} originally formed as a class project benchmark, and a convenient next step after CIFAR-10 and CIFAR-100 since it provides a harder classification task (200 object classes) and we do not have enough resources to run our experiments on ImageNet. Image resolutions are $32 \times 32$ in CIFAR, and $64 \times 64$ in tiny-imagenet. In the analysis sections, we will show that these 3 datasets are quite useful to test the robustness of a hypothesis. We set hyperparameters using validation sets we form by randomly separating $10\%$ of the training set where each class is represented equally. Then, we report test results in the analysis part for which we use the pre-determined test sets for CIFAR-10 and CIFAR-100 and the pre-determined validation set of tiny-imagenet (since we do not have access to the labels of its test set).

\subsection{Models}

We used a well known family of convolutional neural networks: ResNets \cite{he2016deep}. In the original paper, ResNets come with 2 different architectures: a thin ResNet variant for CIFAR-10, and a wide ResNet variant for ImageNet. Although the latter is not specifically designed for the datasets we have, it is essential to test generalization of pruning hypotheses. Moreover, wide ResNets enables using ImageNet pre-trained weights to test the effect of model initialization on pruning. In this study, we employ ResNets \{32, 56, 110\} (\textbf{thin variants}) and ResNet50 (\textbf{wide variant}).

\subsection{$L_1$ penalty} 

This is the hyperparameter $\alpha$ in Eq. \ref{eq:l1}. We tried 4 values: \{$10^{-3}$, $4 * 10^{-4}$, $10^{-4}$, $10^{-5}$\}. Naturally, among these, $10^{-3}$ results in the highest pruning rate and the lowest validation accuracy. $10^{-5}$ produced slightly better classification performance than $10^{-4}$, but with a significantly lower pruning rate. $4 * 10^{-4}$ was able to reach as high as $\approx 96\%$ pruning rate (for example using ResNet110 for CIFAR-10 with weight or spectral radius as the compression mode while preserving a reasonable validation accuracy of $\approx 81\%$) resulting in very sparse models. However, when we set performance of training without weight regularization as baseline, the gap between the validation accuracy of $4 * 10^{-4}$ and the baseline grows as the dataset gets more complex (CIFAR-10 $\rightarrow$ CIFAR-100 $\rightarrow$ tiny-imagenet). Therefore, in our analyses $L_1$ penalty $\alpha = 10^{-4}$.

\begin{equation}
\label{eq:l1}
    L_1 \textnormal{ loss term} = \alpha \sum_{i=0}^n |w_i|
\end{equation}

\subsection{Significance threshold}

\begin{figure*}
\centering
\includegraphics[scale=0.45]{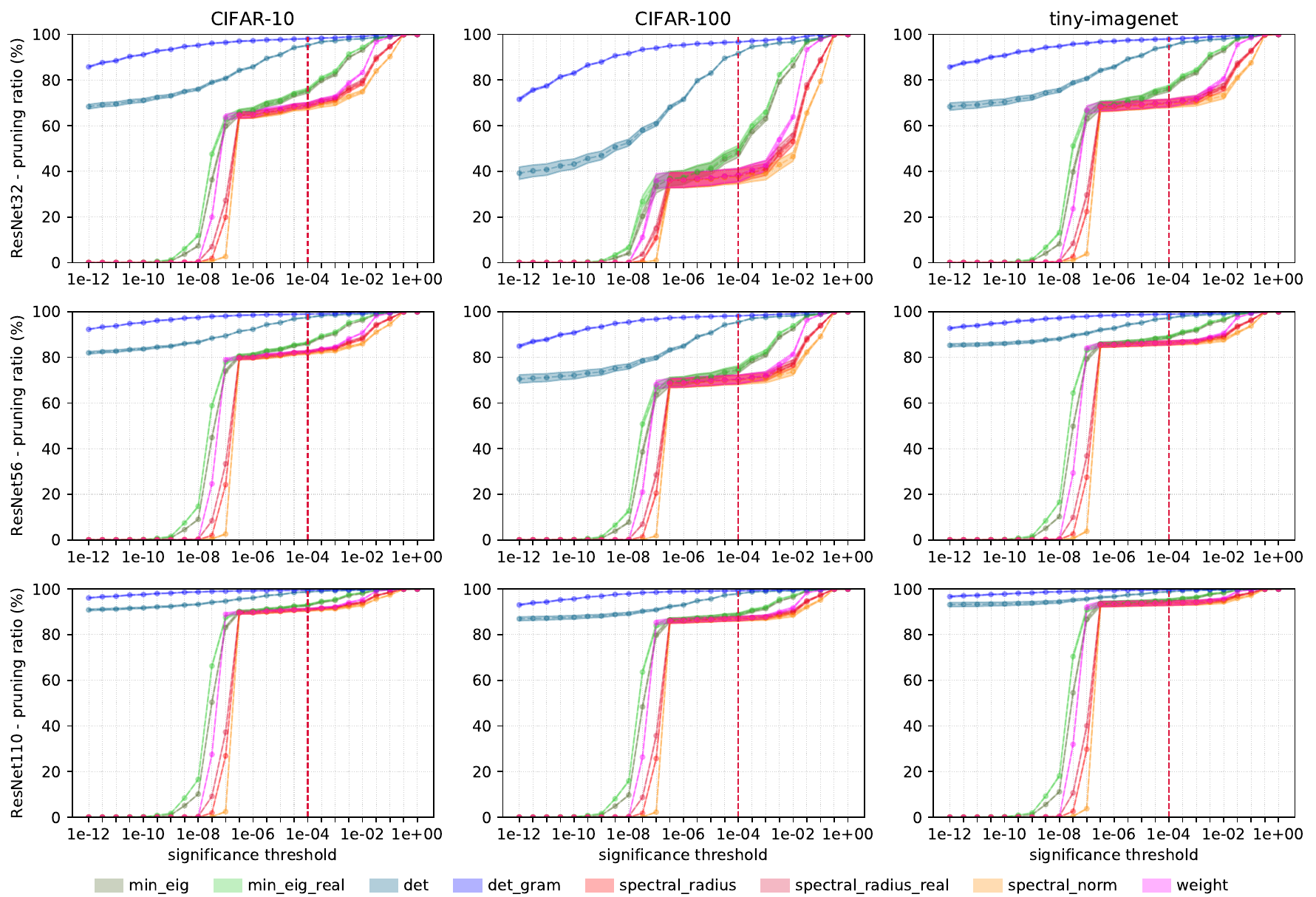}
\caption{Pruning ratio vs. threshold chart for thin ResNets under multiple random initializations.}
\label{fig:ThinMicroResNet_pruning_per_threshold}
\end{figure*}

\begin{figure*}
\centering
\includegraphics[scale=0.45]{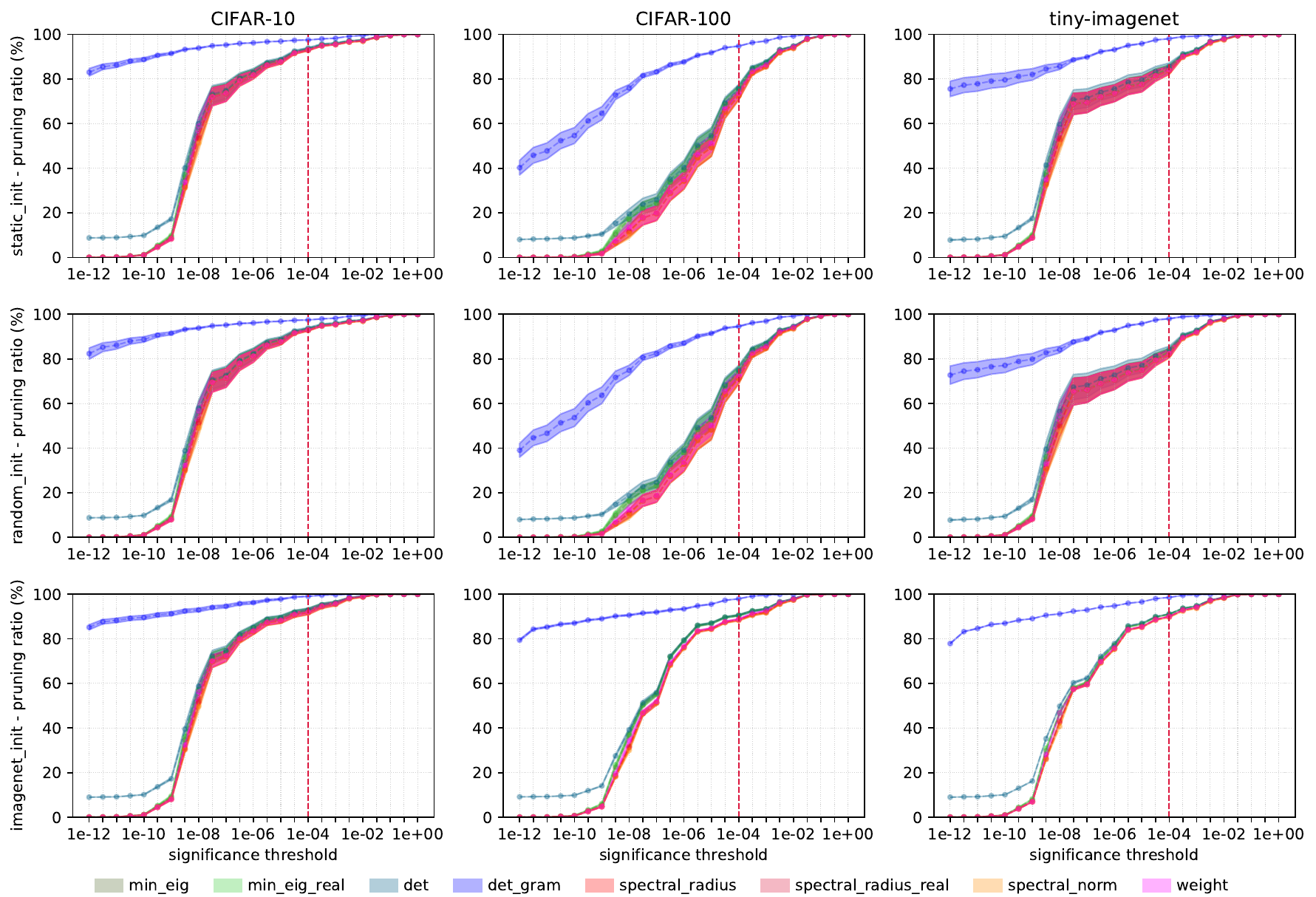}
\caption{Pruning ratio vs. threshold chart for ResNet50 under different initializations.}
\label{fig:MicroResNet50_pruning_per_threshold}
\end{figure*}

Instead of targeting kernels which have significance score of exactly 0, we define a threshold and deem the kernels that have scores under this threshold as insignificant. Hence we name this hyperparameter as significance threshold. In Fig \ref{fig:ThinMicroResNet_pruning_per_threshold} and \ref{fig:MicroResNet50_pruning_per_threshold}, we show the pruning ratio of each compression mode for different significance thresholds to make an informed decision on how to define what values we will take as effectively zero. In Fig \ref{fig:ThinMicroResNet_pruning_per_threshold}, we see similar curves for thin ResNets, and, as expected, the pruning ratio increases as we go from ResNet32 to ResNet110. Moreover, since det and det\_gram modes represent multiplication of eigenvalues/matrices, their curves go significantly above others, so it is important to scale their selected threshold accordingly. 

Another interesting finding is that CIFAR-100 task results in the least pruning ratio compared to CIFAR-10 and tiny-imagenet. In Fig \ref{fig:MicroResNet50_pruning_per_threshold}, the model at hand is ResNet50, which is wider than the ones mentioned before, so it has a different curve. With ResNet50, our goal is to examine the effect of initial weights on pruning. Surprisingly, the pruning ratio curve is similar under different initializations, and it is important to notice that imagenet\_init is substantially different than others since it can actually be used to correctly classify most objects in ImageNet. In addition, in both Figs \ref{fig:ThinMicroResNet_pruning_per_threshold} and \ref{fig:MicroResNet50_pruning_per_threshold}, we see compression modes except det and det\_gram draw quite similar curves where green-ish ones (min\_eig based) go above red-ish ones. As we want to prune the models as much as possible, we did preliminary experiments on a small targeted interval of significance thresholds ($10^{-5} \rightarrow 10^{-3}$). We found $10^{-4}$ to have a good balance between accuracy and pruning ratio. 

In conclusion, in our analysis, we set the significance threshold as:

\begin{itemize}
    \item det: $10^{-12}$ for 3x3 kernels, $10^{-4}$ for 1x1 kernels
    \item det\_gram: $10^{-24}$ for 3x3 kernels, $10^{-8}$ for 1x1 kernels
    \item others: $10^{-4}$
\end{itemize}

\subsection{Statistical significance} 

We repeated each run for 15+ times, and for performance comparison this number goes up to 40+ times simply due to the chronological order of the implementations. Note that each run implies a particular combination of the settings: $f_{\textnormal{settings}}$(model, dataset, initialization mode). For different comparison modes, the training part is shared. Once the training stops, all compression modes branch off. Hence, the experimental results do not have a random parameter space noise for different compression modes.

\noindent \textbf{Other hyperparameters}. We wanted to keep all hyperparameters constant as much as possible to rule out unwanted noise due to variance of hyperparameters for controlled experiments, so unless specifically indicated, a given value for a hyperparameter holds for all settings (models, datasets, etc.):

\begin{itemize}
    \item Optimization algorithm. In this study, we used Adam optimizer \cite{kingma2014adam} due to its fast convergence on the datasets we use.

    \item Learning rate. For ResNets \{32, 56, 110\}, the base learning rate is $10^{-3}$ whereas it is $10^{-4}$ for ResNet50, and for both families it is divided by 10 at epochs 80, 120 and 160. 
    
    \item Batch size. We tried 32, 64, and 128, but we did not found a significant or constant advantage for a particular selection. Therefore, in order to speed-up the experiments, we selected 128 as the mini batch size.
    
    \item Epochs. We monitored the validation accuracy of all models for all datasets, and decided to stop the training at a safe point (validation accuracy curve becomes flat). Hence, the number of epochs is 200, which is an extremely safe point for some cases where learning stabilizes very early on. 
    
    \item Initialization. We have 3 initialization modes:
    \begin{itemize}
        \item random\_init: Models are initialized using He normal initialization \cite{he2015delving} option in Keras \cite{chollet2015keras}.
        
        \item static\_init: We still use He initialization, but we import the same frozen initial weight for each run (we run the experiments multiple times to check statistical significance of the results). This mode provides a control group for random\_init experiments since we do not have a variance in initial conditions here.
        
        \item imagenet\_init: This mode is only available for ResNet50 experiments. Nevertheless, since our compression modes are tightly coupled with the properties of the convolution matrices in models, it is important to test the robustness of our propositions in an unfamiliar parameter space where initial conditions are not random.
    \end{itemize}
    
\end{itemize}

\end{document}